\documentclass[english, 10pt, twocolumn, twoside]{IEEEtran}  
\usepackage{threeparttable}

\usepackage{cite}

\ifCLASSINFOpdf
\else
\fi
\usepackage{graphics}
\usepackage{epstopdf}
\usepackage{subfigure}
\usepackage[cmex10]{amsmath}
\usepackage{amsfonts}
\usepackage{cases}
\usepackage{booktabs}
\usepackage{epsfig}
\usepackage{hyperref}
\usepackage{algorithm}
\usepackage{algorithmic}

\usepackage{array}

\usepackage{multirow}
\usepackage{bigstrut}

\usepackage{amssymb, bm} 
\usepackage{color}

\begin{document}
%
\title{Temporal-Spatial Mapping for Action Recognition}
%
%
\author{Xiaolin~Song,
        Cuiling~Lan,
        Wenjun~Zeng,
        Junliang~Xing,
        Jingyu~Yang,
        and~Xiaoyan~Sun 


}

\maketitle

\begin{abstract}
Deep learning models have enjoyed great success for image related computer vision tasks like image classification and object detection. For video related tasks like human action recognition, however, the advancements are not as significant yet. The main challenge is the lack of effective and efficient models in modeling the rich temporal spatial information in a video. We introduce a simple yet effective operation, termed Temporal-Spatial Mapping (TSM), for capturing the temporal evolution of the frames by jointly analyzing all the frames of a video. We propose a video level 2D feature representation by transforming the convolutional features of all frames to a 2D feature map, referred to as \emph{VideoMap}. With each row being the vectorized feature representation of a frame, the temporal-spatial features are compactly represented, while the temporal dynamic evolution is also well embedded. Based on the VideoMap representation, we further propose a temporal attention model within a shallow convolutional neural network to efficiently exploit the temporal-spatial dynamics. The experiment results show that the proposed scheme achieves the state-of-the-art performance, with 4.2\% accuracy gain over Temporal Segment Network (TSN), a competing baseline method, on the challenging human action benchmark dataset HMDB51.

\end{abstract}


%
\IEEEpeerreviewmaketitle

\section{Introduction}
\label{secIntroduction}

\IEEEPARstart{A}{ction} recognition is an important yet challenging problem in computer vision, with many practical applications such as visual surveillance, human computer interaction, and video analyses \cite{aggarwal2011human}. Recently, deep learning models like Convolutional Neural Networks (CNN) \cite{simonyan2014two,wang2016temporal,zhu2016key,feichtenhofer2016convolutional} and Recurrent Neural Networks (RNN) \cite{yue2015beyond,donahue2015long,sharma2015actionattention,li2016videolstm,wu2016multi,wang2016hierarchical} have been extensively employed to recognize actions in videos. Despite great efforts and rapid developments, the advancements are not as significant as those achieved in image related computer vision tasks such as image classification \cite{krizhevsky2012imagenet,simonyan2014very,he2016deep} and object detection \cite{ren2015faster,liu2016ssd,redmon2016yolo9000}. The main reason is that actions in a video involve not only the spatial information of each frame, but also its temporal evolution. Exploring this rich temporal-spatial information requires the deep learning model to be equipped with more parameters, trained with more video samples, and most importantly formulated with more effective architecture.

Previous attempts to address action recognition include the two-stream ConvNets \cite{simonyan2014two,wang2016temporal,feichtenhofer2016convolutional,diba2017deeptemporal}, 2D ConvNets followed by Long Short-Term Memory (LSTM) networks \cite{donahue2015long,yue2015beyond,li2016videolstm,wu2016multi,wang2016hierarchical}, 3D ConvNets \cite{ji20133d,tran2015learning}, and many others \cite{sharma2015actionattention,li2016action,wang2017spatiotemporal}. These models are continually pushing forward the state-of-the-art performances of action recognition. Most of these models, however, suffer from limitations such as lack of joint temporal-spatial learning \cite{simonyan2014two,wang2016temporal,feichtenhofer2016convolutional,diba2017deeptemporal}, difficulties in model training \cite{donahue2015long,yue2015beyond,li2016videolstm,ji20133d,tran2015learning}. Moreover, limited by the designs, most of those approaches cannot leverage more dense frames for obtaining further gains even though more frames can provide more temporal-spatial information \cite{wang2016temporal,diba2016deep,wang2013action}. That is because the statistics of the features/scores of frames are utilized to get the final prediction \cite{wang2016temporal,diba2016deep}. Frames above a certain number (\emph{e.g.}, 25 frames) help very little once they are enough to estimate the statistics. To overcome these limitations, we need a network architecture which jointly and effectively learns the temporal-spatial feature representations and is capable of exploring the information of dense frames.

\begin{figure}[t]
  \begin{center}
   \includegraphics[width=1.0\linewidth]{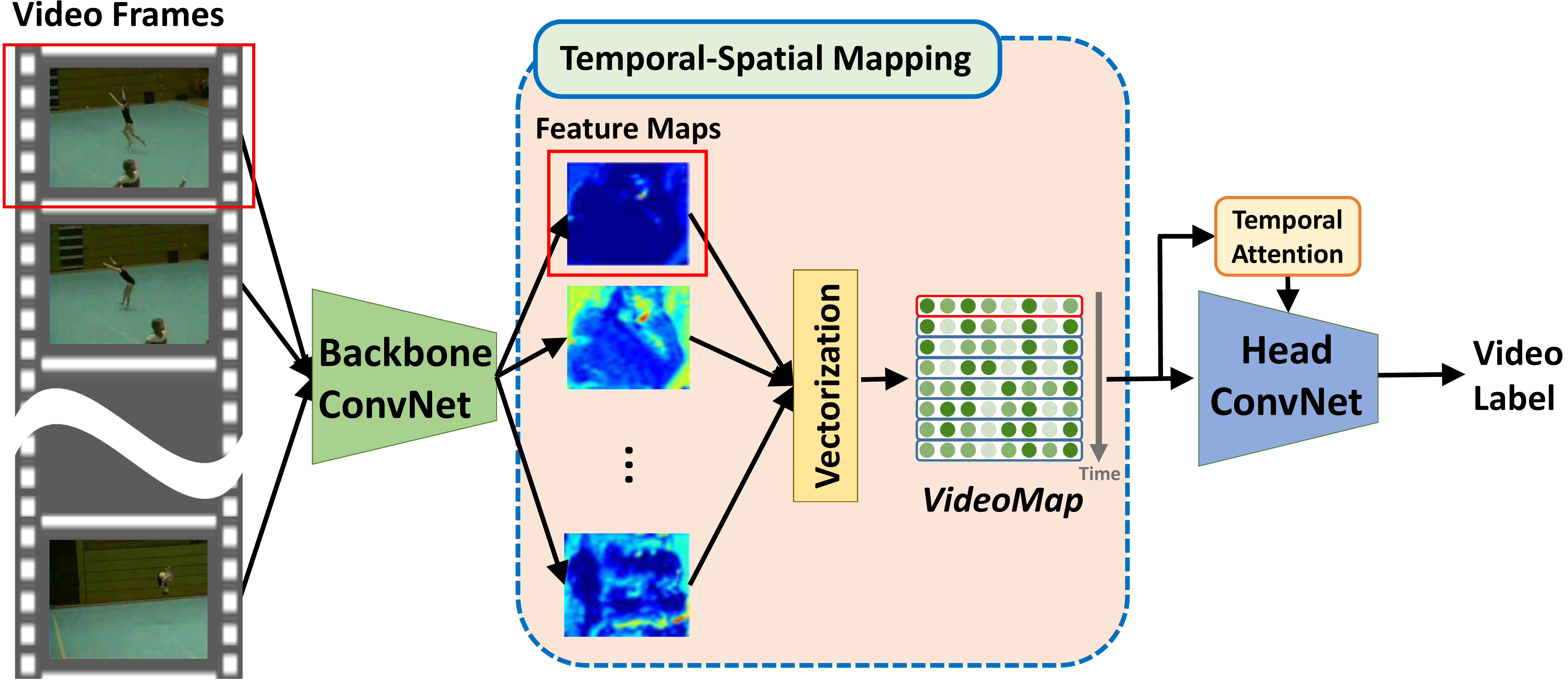}
  \end{center}
  \vspace{-5mm}
  \caption{Overview of our network structure. The Temporal-Spatial Mapping operation enables the effective joint temporal-spatial modeling by representing the temporal spatial features of an entire video by a 2D \emph{VideoMap}. A temporal attention model in a head ConvNet further transforms the  \emph{VideoMap} to a compact video-level feature embedding for classification.}
  \label{fig:TSM-operation}
  \vspace{-3mm}
\end{figure}

To this end, we present a simple yet effective operation, \emph{i.e.}, Temporal-Spatial Mapping (TSM), for joint temporal-spatial feature modeling. We represent the temporal-spatial features of an entire video compactly by a \emph{VideoMap}, which is a row-wise layout of the per-frame vectorized ConvNet features as illustrated in the middle of Figure \ref{fig:TSM-operation}. This enables the ``seeing" of dense frames at a glance and thus performing effective joint temporal-spatial analyses. The proposed TSM operation is general and can be used after any convolutional features for video-level temporal-spatial feature learning.

To deploy this TSM operation for action recognition, we first train a backbone 2D ConvNet model to extract convlutional features for each frame of a video sequence, then the TSM operation is performed on the features to generate the temporal-spatial \emph{VideoMap}, which naturally encodes the temporal-spatial information in 2D feature map. Based on the compact \emph{VideoMap} representation, we further propose a temporal attention model within a head ConvNet to extract effective video-level feature embeddings to predict the final action categories. Experiment results on two large benchmarks, HMDB51 and UCF101, demonstrate the effectiveness of the proposed network architecture and its state-of-the-art performances.

To summarize, the main contributions of this work are three-fold:
\begin{itemize}
\item We propose a simple yet effective operation, Temporal-Spatial Mapping, for jointly embedding the temporal-spatial information of a video from per-frame features into a compact feature map, \emph{i.e.}, \emph{VideoMap}. The proposed operation is general and can be applied to any CNN features to explore temporal dynamics. \emph{VideoMap} representation provides a way to leverage dense frames for enhancing the performance.
\item We propose a temporal attention model within a head ConvNet to further transform the temporal-spatial \emph{VideoMap} to a more compact and effective video-level feature representation for classification, which can better exploit the temporal-spatial dynamics.
\item We present a deep architecture for action recognition which achieves significant performance improvement on the HMDB51 dataset. The source code and trained model will be released to facilitate the research in action recognition.
\end{itemize}

\section{Related Work}
\label{secRelWork}
Motivated by the outstanding performance of deep neural networks on image classification and detection, more and more works have extended CNN-based or CNN-LSTM-based architectures for video analysis.

The two-stream ConvNets approaches \cite{simonyan2014two,wang2016temporal} train two separate 2D ConvNets for both appearance in still images and stacks of optical flow, from several sparsely sampled frames. The temporal stream takes the short-term temporal information into account by means of optical flow and achieves superior performance than the spatial stream (still images). The Temporal Segment Networks (TSN) \cite{wang2016temporal} combine a sparse temporal sampling strategy and video-level supervision in training to explore temporal structure. Three frames which are randomly selected from three equally divided segments are jointly trained with frames combined by average pooling. However, these approaches involve temporal information by simply averaging/multiplying the scores across frames for the video level prediction. Such approaches still cannot accurately model the temporal dynamics \cite{simonyan2014two,wang2016temporal,diba2016deep}. Some aggregation techniques like VLAD \cite{Jegou2012Aggregating}, Fisher Vector \cite{Nchez2013Image} and dictionary learning \cite{xu2017two-stream} have been used in action recognition for aggregating features of frames  \cite{Girdhar2017ActionVLAD,xu2015discriminative,peng2014action}. VLAD-based methods like ActionVLAD \cite{Girdhar2017ActionVLAD} provide solutions to perform spatio-temporal aggregation of a set of action primitives. However, they do not model the time order of frames.

To extend convolution operations from 2D image to 3D video, the 3D ConvNets \cite{tran2015learning} directly operates on the video for spatio-temporal feature learning by replacing 2D filters with 3D ones. So far, such approach however has shown limited benefit, probably due to the lack of training data, high complexity of training 3D convolution kernels, and not exploiting the optical flow stream explicitly. To reduce the number of parameters of 3D ConvNets, Sun \emph{et al.} \cite{sun2015human} propose to factorize the original 3D convolution kernel learning as a sequential process of learning 2D spatial kernels followed by learning 1D temporal kernels. However, both approaches model the temporal dynamics by averaging the activations of the sub-clips of a video. There is still a lack of a global modeling of the action among the video sub-clips. To capture the relationships among the video sub-clips, the temporal linear encoding (TLE) \cite{diba2016deep} encodes the aggregated information of $K ~(\emph{i.e.}, K=3)$ clips/frames into a video feature representation by performing element-wise multiplication of the features of the clips/frames. However, for a long video, the sampling of $K$ clips/frames and the aggregation of them by multiplication results in information loss. Since the statistics of clips/frames are explored rather than their details, like TSN \cite{wang2016temporal}, the performance increases little when more frames are used.

In \cite{yue2015beyond,donahue2015long,sharma2015actionattention,li2016videolstm}, LSTM is utilized to explore the temporal evolution of the per-frame CNN features across a video. Shi \emph{et al.} \cite{shi2017sequential} propose a sequential Deep Trajectory Descriptor (sDTD) to model long-term motion information in video and employe a three-stream CNN-LSTM architecture for action recognition. Wang \emph{et al.} \cite{wang2018two-stream} propose two-stream 3D ConvNets Fusion to recognize actions of arbitrary size and length by using spatial temporal pyramid pooling (STPP) with a LSTM or CNN model to extract multi-size descriptions and learn global representation for each input video. Li \emph{et al.} \cite{LiUnified} propose a unified Spatio-Temporal Attention Networks (STAN) using attention neural cell (\emph{AttCell}) based on CNN-LSTM architecture to estimate attention on both spatial and temporal locations in a video. Compared with image-based approaches \cite{simonyan2014two,wang2016temporal}, LSTM-based approaches go one step further which can model the temporal dynamics of a video. However, applying the LSTM models to video based action recognition has so far only achieved similar performance as temporal pooling \cite{yue2015beyond}, likely attributed to the rigid structure of LSTM and the difficulties in training.


In this paper, we propose a general approach of temporal-spatial mapping to facilitate the joint analysis of the dense frames/clips of a video, with the time order information embedded in the mapped \emph{VideoMap}. Our approach provides an efficient way to explore the details of dense frames, enabling performance improvement. 

\section{Temporal-Spatial Mapping Operation}
\begin{figure*}[t]
 \centering\includegraphics[width=\textwidth]{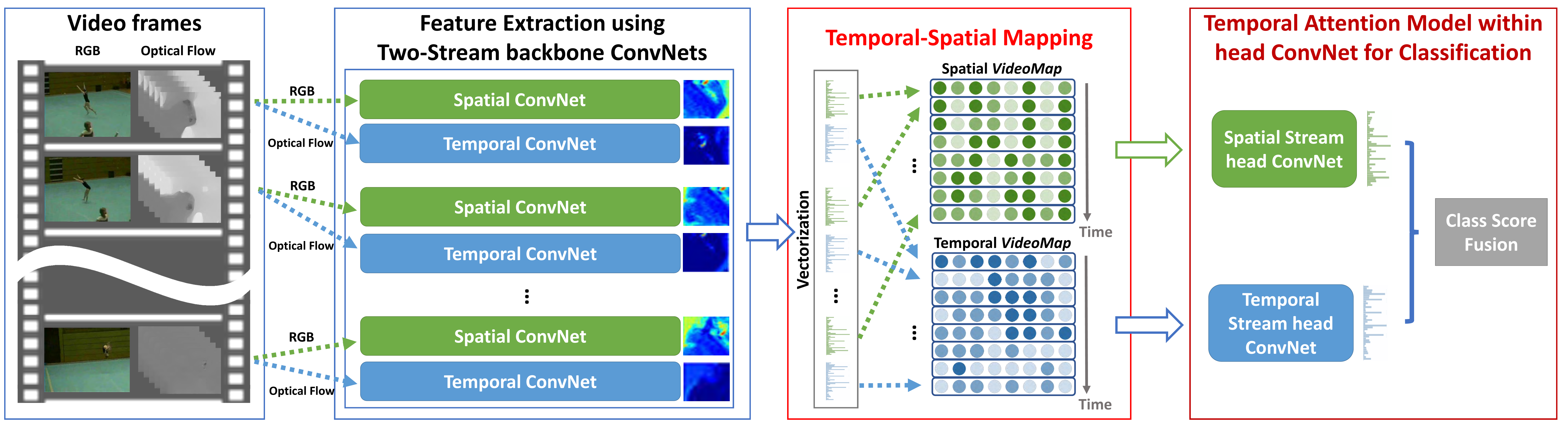}
 \vspace{-5mm}
 \caption{The overall framework with our Temporal-Spatial Mapping operation followed by a head ConvNet for action recognition. Two-stream ConvNets extract features on each frame for the spatial stream (RGB) and temporal stream (Optical flow), respectively. The vectorized feature vectors of the sequential frames form a \emph{VideoMap} for temporal-spatial representation. A head ConvNet with temporal attention makes action classification based on the \emph{VideoMap}. Finally the class scores of the \emph{VideoMaps} from two streams are fused to produce the video-level prediction.}
 \label{fig:arch}
 \vspace{-3mm}
\end{figure*}
In a video, besides the spatial information in each image, the temporal evolution provides vital information for identifying an action. It is not trivial to find a video representation that encodes the dense frames together to facilitate the joint analysis of the entire video. The traditional powerful approach iDTs \cite{wang2013action} densely samples feature points in video frames and uses optical flow to track them to yield a good video representation. For action recognition from video, most deep learning based approaches which have good performance \cite{simonyan2014two,wang2016temporal,diba2017deeptemporal} are still image based where they infer the final results by averaging/multiplying the scores/features of frames. The proposed Temporal-Spatial Mapping provides a way to jointly consider the dense sequential frames for inferencing the video label.



\begin{figure}[t]
\begin{center}
\includegraphics[width=0.8\linewidth]{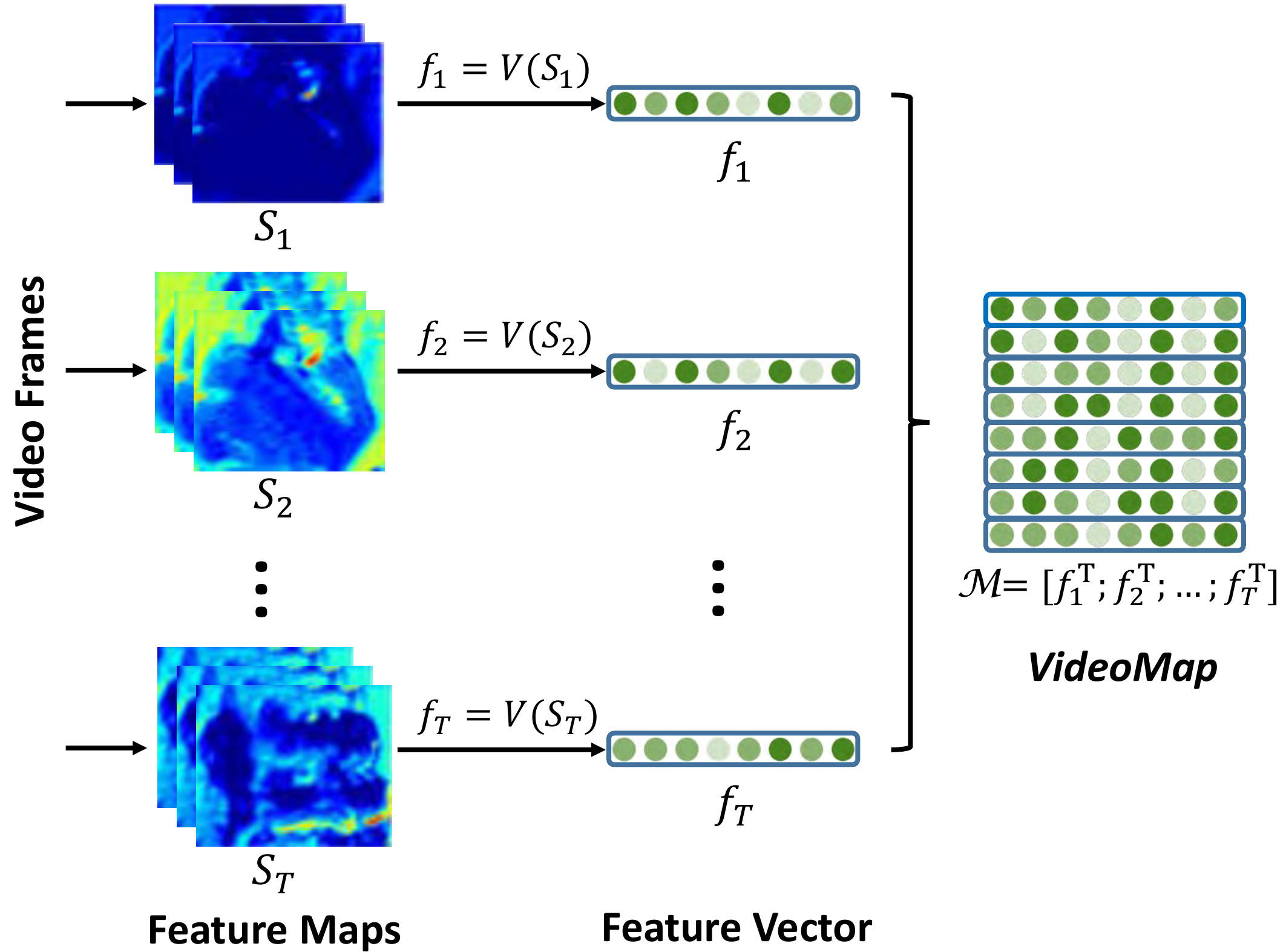}
\end{center}
\vspace{-7mm}
\begin{center}
\caption[width=0.95\linewidth]{Illustration of the proposed Temporal-Spatial Mapping operation, which transforms a sequence of feature maps into a compact \emph{VideoMap}.}
\end{center}
\label{fig:TSM-operation-detail}
\end{figure}

Figure~\ref{fig:TSM-operation-detail} shows the process. For each frame of a video sequence, ConvNet generates feature maps at each convolutional layer, where features of higher abstraction are captured by higher layers \cite{szegedy2015going}. Consider the output features of 2D ConvNet for $T$ frames from a video. The features of the $k^{th}$ frame can be a feature vector of $L-$dimension, \emph{i.e.}, $f_k \in \mathbb{R}^L$, \emph{e.g.}, the output of the global pooling layer of the TSN~\cite{wang2016temporal}. They can be feature maps with high dimensions, \emph{i.e.}, $S_k \in \mathbb{R}^{h\times w \times c}$, where $h$, $w$ and $c$ denote the height, width, and number of channels of the feature maps, \emph{e.g.}, the inception layer output of TSN \cite{wang2016temporal}. The high dimensionality of feature maps makes it challenging to jointly analyze the dense frames of a video. Thus, in that case, a spatial vectorization function $V: S_k \rightarrow f_k$, is used to encode the feature maps to a low dimension vector of fixed length, \emph{i.e.}, $f_k = V(S_k)$, where $f_k \in \mathbb{R}^L$.  Then, we layout the feature vector of each frame as a row with the row identity corresponding to the time order of the frames to create a two-dimensional temporal-spatial map, \emph{i.e.}, \emph{VideoMap} as
\begin{equation}
\mathcal{M} = [f_1^{\top}; f_2^{\top}; \cdots; f_T^{\top}] \in \mathbb{R}^{T \times L}.
\end{equation}
The width of the map is equal to the total number of frames while the height is equal to the dimension of the feature vector. The map has embedded both the temporal and spatial information. This makes it possible to have a global observation of a video sequence and facilitates the exploration of the temporal dynamics.

This TSM operation has the following characteristics and advantages.
\begin{itemize}
 \setlength{\itemsep}{0pt}
 \setlength{\parsep}{0pt}
 \setlength{\parskip}{0pt}
 \item This TSM operation is a general operation which can be applied to the feature maps/features from ConvNets for encoding the temporal-spatial dynamics from a sequence of frames.
 \item This TSM operation for obtaining a \emph{VideoMap} can maintain the time order information of the dense frames, which helps distinguishing action categories related with occurrence order, \emph{e.g.}, ``stand up'' versus ``sit down''.
 \item This TSM operation is simple yet effective. It does not involve complicated operations.
\end{itemize}

\noindent\textbf{Discussions.} Here, we discuss the relation and differences of our method with several classical methods. We aim to explore the temporal dynamics from the dense frames of a video and jointly make a decision from them.

\emph{Two-stream based ConvNets:} Our framework is compatible with the two-stream ConvNet based approaches \cite{simonyan2014two,wang2016temporal}. TSN \cite{wang2016temporal} uses three temporal segmented frames to explore the long-range temporal structure in training. During testing, the scores from $N (N=25)$ frames are averaged to finally predict the action. However, the temporal dynamics are only weakly explored by the simple averaging of a few frames without considering the time order. In contrast, our temporal-spatial mapping to a \emph{VideoMap} enables the joint exploration of many frames with time order retained.

\emph{3D Convolution:}~3D convolution provides an elegant framework for exploring the spatial and temporal dynamics \cite{ji20133d,tran2015learning}. Without the practical constraints, such as on memory, labeled data, computational resource, it is expected to achieve excellent performance. However, such approaches so far have not demonstrated satisfactory performance in practice due to the difficult to train. In practice, 3D Convolution only covers a short range of the sequence for each input video sub-clip (\emph{e.g.}, 5-7 frames in \cite{ji20133d}, 16 frames in C3D \cite{tran2015learning}). The scores of sub-clips are averaged to get the final prediction. The temporal dynamics among clips are not well explored by the simple averaging and the time order information among the sub-clips is lost.

\emph{CNN+LSTM:} To tackle the not well solved temporal dynamic modeling problem, some works \cite{yue2015beyond,donahue2015long,sharma2015actionattention,li2016videolstm} use the Recurrent Neural Network with Long-short Term Memory (LSTM) to model the temporal evolution. The RNN structure facilitates the exploration of temporal dynamics from the dense frames with time order considered. However, it has only achieved similar performance as temporal pooling \cite{yue2015beyond}. This might be attributed to the difficulty in training with the gradient vanishing for the long videos.

In contrast, we leverage the temporal-spatial mapping to obtain a \emph{VideoMap} which embeds the information of temporal dynamics and time order. It facilitates the joint exploration of the dense frames of a video for a global decision.

\section{Temporal Attention within Head ConvNet}
\label{ssecAttention}

Motivated by the success of using ConvNet for feature extraction in image classification, which jointly explores cross-pixel correlations, we leverage a convolution neural network to jointly explore the cross-frame dynamics.

\begin{figure}[t]
\begin{center}
\includegraphics[width=0.95\linewidth]{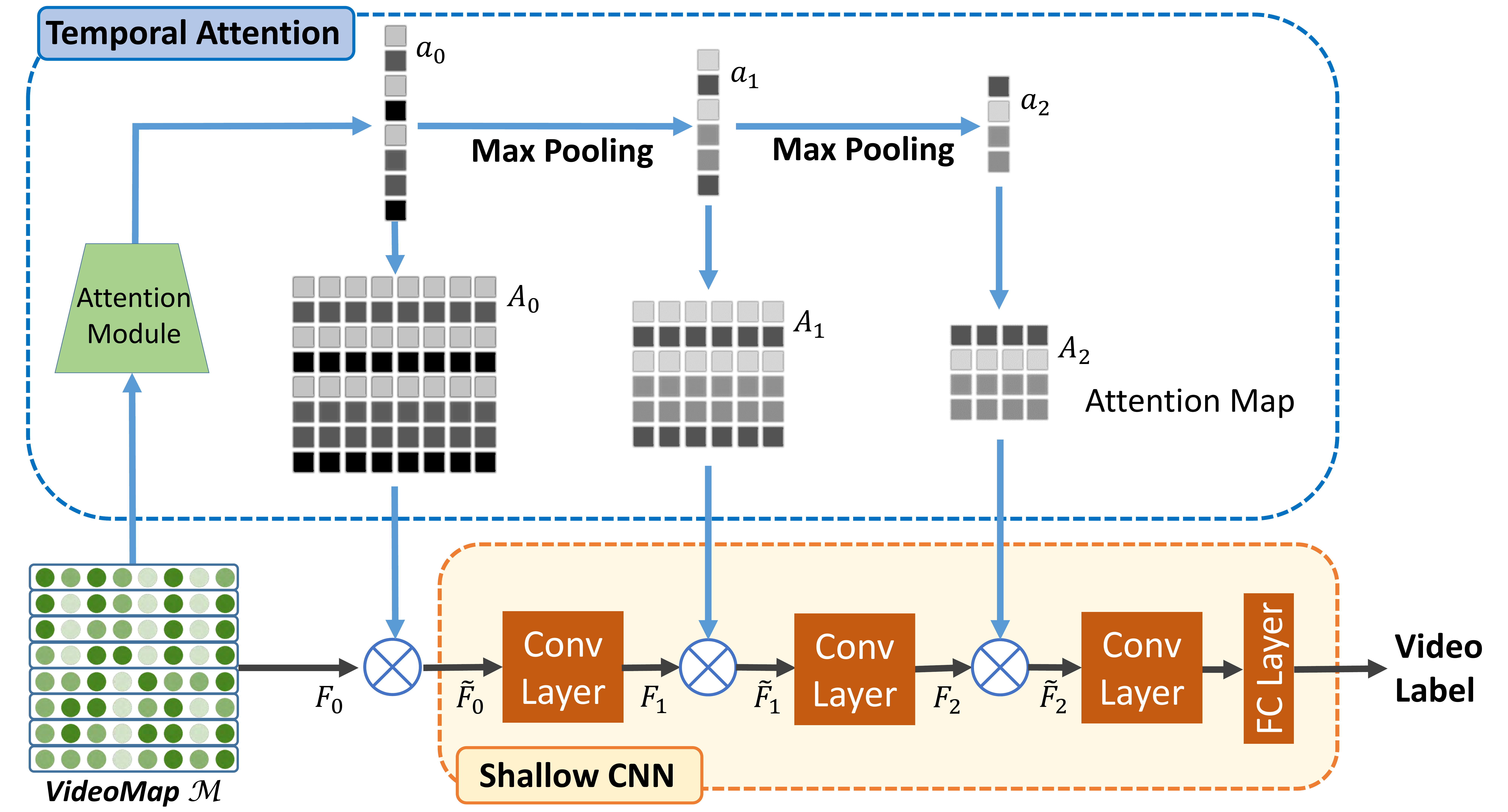}
\end{center}
\vspace{-7mm}
\begin{center}
\caption[width=0.95\linewidth]{Temporal attention model within a head ConvNet for action recognition from \emph{VideoMap}.}
\end{center}
\label{fig:shallow-cnn}
\vspace{-4mm}
\end{figure}

With the \emph{VideoMap} as input, we design a temporal attention model within a head ConvNet for video level feature extraction and action recognition. Figure \ref{fig:shallow-cnn} shows this network structure, which consists of a shallow ConvNet and a temporal attention module. Note that we refer to this shallow ConvNet as head ConvNet since it is the last sub-network specific to the task \cite{he2017mask}. The responses from the temporal attention module are incorporated into the head ConvNet to adjust the importance level of temporal features. Cross-entropy as used in \cite{wang2016temporal} is taken as the video level loss function.

To recognize the action class in a video, the importance level of different frames differs. Some frames are more likely to be irrelevant or less relevant to the action category and may hurt the final performance by introducing noise. Some other frames are more relevant to the action category. Take the action of handshake as an example, the frames with two people approaching are less relevant to the action, which could be shared by other action types, while the frames with two people's hands holding together give more discriminative information. Therefore, we introduce a temporal attention model for learning and determining the importance levels.

As illustrated in Figure \ref{fig:shallow-cnn}, the feature maps $\tilde{F_{i}}$ of the $i^{th}$ layer of the head ConvNet after enforcing the temporal attention are described as:

\begin{equation}
\tilde{F_{i}} = A_i(\mathcal{M}) \circ F_{i}(\mathcal{M}) , 
\label{attention}
\end{equation}

\noindent where $F_{i}(\mathcal{M})$ denotes the output feature maps of the $i^{th}$ layer of the head ConvNet, $A_i(\mathcal{M})$ is the attention map from the attention model, $\circ$ denotes entrywise product. Here $A_i(\mathcal{M}) = a_i(\mathcal{M}) \otimes \mathbf{1}$, where $\otimes$ is the Kronecker product and $\mathbf{1}$ denotes the all-ones vector, $a_i(\mathcal{M})$ is the learned attention vector with the dimension being related with the total number of frames. In other words, $A_i(\mathcal{M})$ is the column-wise repeat of $a_i(\mathcal{M})$. Note that for the $0^{th}$ layer, $F_{0}(\mathcal{M}) = \mathcal{M}$, $a_0(\mathcal{M}) \in \mathbb{R}^{\top}$. $a_i(\mathcal{M})$ is the max pooling result of $a_{i-1}(\mathcal{M})$ with a stride of 2. The detailed network designs will be described in Section \ref{secExperiments}.

\section{Overall Framework}


We take the two-stream based ConvNets as our backbone ConvNets and embed the proposed TSM operation followed by a head ConvNet with temporal attention model, for the video-level classification. Figure~\ref{fig:arch} shows the overall flowchart of our final framework.

Temporal Segment Networks (TSN) \cite{wang2016temporal} with BN-Inception structure \cite{ioffe2015batch} provides superior performance on both the spatial and temporal streams. We take the TSN as our backbone ConvNets for frame feature extraction. The network contains two streams: spatial stream with RGB image as input, and temporal stream with optical flow as input. The results from the two streams are fused to predict the video label.

Without loss of generality, we take the spatial stream as example to describe our overall network structure. The temporal stream acts similarly. For successive of frames in a video, the backbone spatial ConvNet outputs feature maps for each frame. With the feature maps of each frame vectorized to a feature vector, the feature vectors of the successive frames are arranged row-by-row to form a \emph{VideoMap}. The \emph{VideoMap} goes through the head ConvNet with temporal attention and the class scores are generated. The Temporal-Spatial Mapping operation permits the end to end training of the entire network. Due to memory constraints, in practice, we train the networks in two stages. In the first stage, we train the backbone ConvNets. Then we train the the head ConvNet for \emph{VideoMap} classification.

\section{Experiments}
\label{secExperiments}
We validate the effectiveness of the proposed framework on two benchmark datasets. We first describe the datasets and implementation details. Then we study the effects of different factors in our network. Finally, we compare our approach with many state-of-the-art approaches. 

\subsection{Datasets}

We conduct experiments on two popular human action recognition datasets, namely HMDB51 \cite{kuehne2011hmdb} and UCF101 \cite{soomro2012ucf101}. The HMDB51 dataset is a very challenging dataset with higher intra-class variations and smaller inter-class variations. The videos are collected from movies and a variety of YouTube consumer videos. This dataset consists of 6,766 video clips from 51 action categories, with each category containing at least 100 clips. In each split, each action class has around 70 clips for training and 30 clips for testing. The UCF101 dataset consists of 13320 video clips in 101 categories. This dataset provides large diversity in terms of actions, variations in background, illumination, camera motion and viewpoints, as well as object appearance, scale and pose.

\subsection{Implementation Details}
\noindent\textbf{Two-stream Backbone ConvNets.}
We pre-train our backbone network of the TSN the same way as reported in \cite{wang2016temporal}. The size of input images or optical flow stacks is fixed at $256\times340$. In order to avoid over-fitting, we perform data augmentation, by cropping images with the width and height chosen from four different sizes, 256, 224, 192 and 168, from five spatial locations of the full image, \emph{i.e.}, one center and four corners. These cropped regions will be resized to $224\times224$ for feature extraction. Note that, to maintain spatial consistency across a video, the cropped size and location are the same across a video sample.


\noindent\textbf{Temporal-Spatial Mapping and head ConvNet.} For the $k^{th}$ frame, the top inception module of the TSN outputs feature maps $S_k$ of size $h\times w\times c$, where $h=7$,$w=7$, and $c=1024$. An average pooling on each channel vectorizes them to a 1024-dimension vector $f_k$. For $T$ sequential frames, a \emph{VideoMap} $\mathcal{M} = [f_1^{\top}; f_2^{\top}; \cdots; f_T^{\top}]$ is obtained.

Video level feature learning and classification based on the \emph{VideoMap} is performed using our head ConvNet with temporal attention. We build the head ConvNet by stacking three convolution blocks. In each block, it consists of a convolutional layer with 5$\times$5 kernels, a ReLU layer, followed by a pooling layer of 3$\times$3 sized kernel of stride 2. We construct the temporal attention module by two such convolution blocks followed by a fully connected layer which outputs a $T$ dimensional vector, representing the frame-wise attention responses for a video. Since the GPU memory is limited, we set the mini-batch to have 128 \emph{VideoMaps}. We set the initial base learning rate to 0.01 and decrease it by a factor of 10 every 10,000 iterations. We stop the training process after 100 epochs.


\noindent\textbf{Vectorization of Feature Maps.} For the high level convolution feature maps, the average pooling (\emph{e.g.}, BN-Inception \cite{wang2016temporal}) or full connection (\emph{e.g.}, AlexNet \cite{simonyan2014two}) of the feature maps has been done in the ConvNets structure, to convert them to a low dimensional feature vector. Therefore, we directly utilize such feature vectors from the sequential frames to form a \emph{VideoMap}.

Given the feature maps of high dimension from a ConvNet layer, if the feature maps are not already transformed to a feature vector, a module for vectorization of the feature maps is needed to convert the feature maps to a low dimensional feature vector representation. There are many ways to perform to perform the vectorization, \emph{e.g.}, leveraging a ConvNet to encode the feature maps to a feature vector.



\subsection{Ablation Study}

In this subsection, we will analyze the effectiveness of the proposed Temporal-Spatial Mapping component, discuss the design of the head ConvNet, analyze the effectiveness of the temporal attention module, and the influence of the height of the \emph{VideoMap} (\emph{i.e.}, density of temporal sampling), respectively.

\noindent\textbf{Effectiveness of Temporal-Spatial Mapping.} Aggregation of single-frame features of dense frames to a \emph{VideoMap} provides the opportunity to jointly explore the temporal-spatial dynamics at the video level. We show the performance of our scheme in comparison with the baseline scheme on the HMDB51 dataset (Split 1) in Table \ref{tab:twostream}.

In Table \ref{tab:twostream}, ``TSN~\cite{wang2016temporal}" denotes the results of the TSN approach \cite{wang2016temporal} with the BN-Inception structure, serving as our baseline scheme. We take its ConvNets as our backbone network to extract frame level features. ``TSN+TSM~(Ours)" denotes our scheme with the Temporal-Spatial Mapping operation and a head ConvNet for the joint temporal dynamics exploration from densely sampled frames. We can see that our scheme achieves 1.6\% after the two-stream fusion. Note that in Table \ref{tab:twostream} and Table \ref{tab:vscnn}, TSN results are obtained from the original TSN model but ran based on a new Caffe version on Windows. \footnote{The results reported in website (\url{http://yjxiong.me/others/tsn/\#exp}, spatial stream, temporal stream and the final fusion are 54.4\%, 62.4\% and 69.5\%, respectively) of the original TSN are a little different from our ran results due to the difference of Caffe version on Windows. }

We also evaluate the performance when using another backbone network, which takes two-stream ConvNets with VGG-16 network structure \cite{feichtenhofer2016convolutional} as the frame-level feature extractor. Similarly, TSM brings 1.8\% improvement in accuracy.

\begin{table}[t]
 \caption{Comparison with two-stream based networks in accuracy (\%) on the HMDB51 dataset (Split 1). Here ``Two-stream'' \cite{simonyan2014two} uses VGG-16 as the network structure, and ``TSN \cite{wang2016temporal}'' uses BN-Inception. ``TSN+TSM(Ours)" is our scheme with the Temporal-spatial Mapping (TSM) followed by a head ConvNet.}
 \label{tab:twostream}
 \begin{center}
   \fontsize{9pt}{10pt}\selectfont\centering
   \begin{tabular}{|l|c|c|c|}
     \hline
     Method & RGB & Optical Flow & Fusion \\
     \hline\hline
     Two-stream \cite{feichtenhofer2016convolutional} & 42.2 & 55.0 & 58.5 \\
     Two-stream+TSM~(Ours) & \textbf{43.1} & \textbf{56.2} & \textbf{60.3} \\
     \hline
     TSN \cite{wang2016temporal} & 54.6 & 62.6 & 70.8 \\
     TSN+TSM~(Ours) & \textbf{55.0} & \textbf{63.1} & \textbf{72.4} \\
     \hline
   \end{tabular}
 \end{center}
\vspace{-4mm}
\end{table}

\begin{figure*}[t]
	\includegraphics[width=\linewidth]{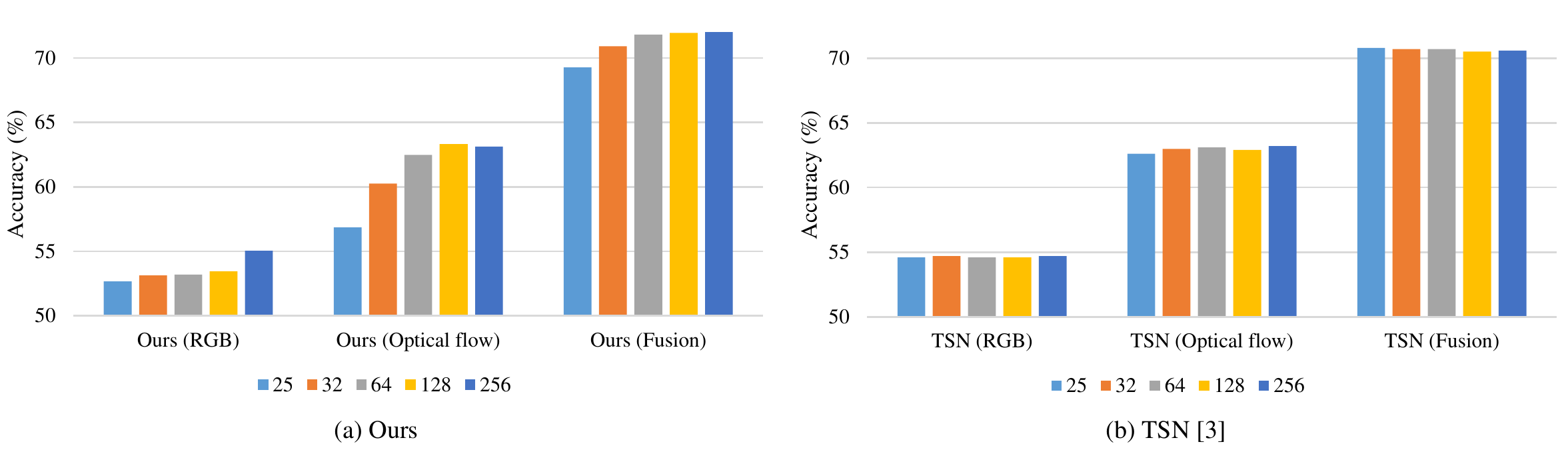}
	\vspace{-5mm}
	\caption{Comparison of performance (accuracy \%) at different frame sampling densities (number of frames: 25, 32, 64, 128, 256, respectively) for our method and the TSN on the HMDB51 dataset (Split 1). (a) Ours; (b) TSN~\cite{wang2016temporal}. In our scheme, the performance increases as the sampling density increases. In TSN, however, the performance does not increase as the sampling density increases.}
	\label{fig:redun}
	\vspace{-3mm}
\end{figure*}

\begin{figure*}
	\begin{center}
		\includegraphics[width=0.9\linewidth]{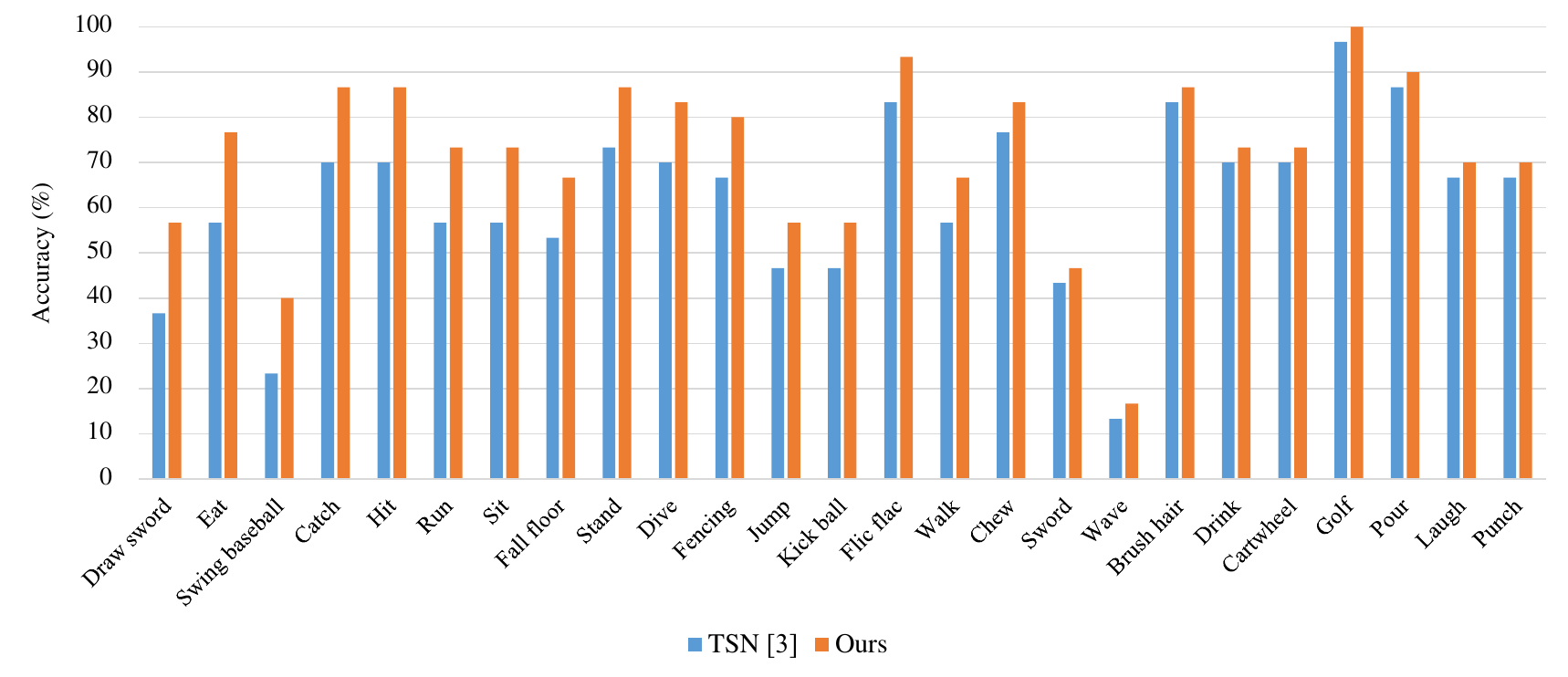}
		\vspace{-3mm}
		\caption{Comparisons of Accuracy (\%) for the top-25 classes on the HMDB51 dataset (Split 1) between our approach and the TSN model. Our approach consistently outperforms the TSN model.}
		\label{fig:histgram-gainclasses}
	\end{center}
\end{figure*}

\begin{figure*}
	\begin{center}
		\includegraphics[width=1.0\linewidth]{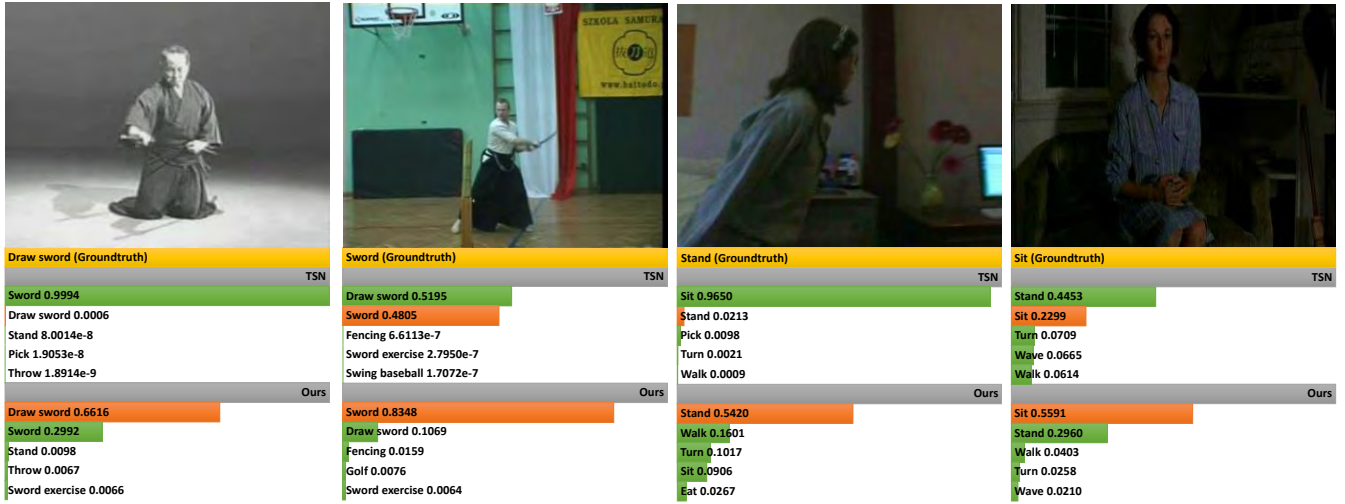}
		\vspace{-3mm}
		\caption{Examples that our scheme succeeds in recognizing the action while TSN \cite{wang2016temporal} fails. Our scheme perform well mainly due to the Temporal-Spatial Mapping operation enables the joint exploration of temporal frames and the consideration of time order.}
		\label{fig:examples}
	\end{center}
\end{figure*}


\begin{figure*}
	\begin{center}
		\includegraphics[width=0.93\linewidth]{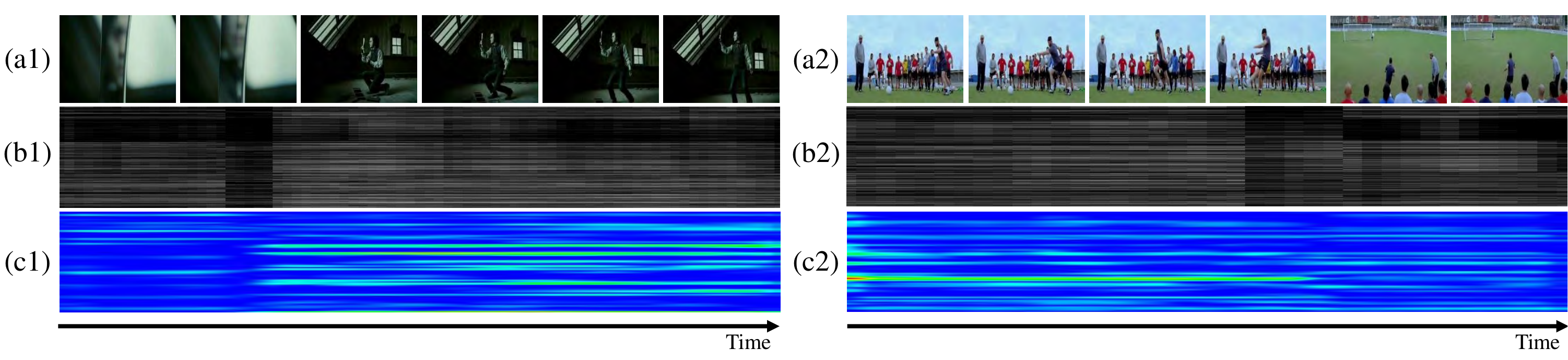}
	\end{center}
	\vspace{-3mm}
	\caption{Visualizations on a video of action ``stand'' ((a1)--(c1)) and ``kick ball'' ((a2)--(c2)). (a1)(a2) Video frames over time (only some frames are shown). (b1)(b2) VideoMap. (c1)(c2) Visualization from our ConvNet of the Conv-3 layer using approach Grad-CAM \cite{Selvaraju2016Grad}. Note that images in (b1)(c1)(b2)(c2) are resized and horizontal axis denotes the time here. We can see that Grad-CAM map presents higher responses over temporal segments corresponding to the frames when the persons are doing the corresponding actions.}
	\label{fig:visualization}
\end{figure*}

\noindent\textbf{Comparisons on Network Designs.} We have designed a head ConvNet for encoding the \emph{VideoMap} for action recognition. The purpose of this network is to explore temporal dynamics among frames to learn efficient video feature representation. For the head ConvNet with \emph{VideoMap} as input, we have tried three network structures: AlexNet, our designed 3-layer ConvNet, and 1-layer ConvNet. In addition, one alternative way for temporal modeling of the feature vectors is to use the Recurrent Neuron Network with LSTM. However, RNN suffers from the gradient vanishing problem even though LSTM suffers less than RNN. We show the results of these designs in Table \ref{tab:vscnn} with experiments conducted on the HMDB51 dataset (Split 1). We can see that our 3-layer ConvNet (\emph{TSN+TSM~(3-layer ConvNet)}) achieves much superior performance than the LSTM based network. The performance of AlexNet is inferior to the 3-layer ConvNet since a deeper network with more parameters is prone to over-fitting. The 1-layer ConvNet does not converge in modeling the temporal-spatial dynamics.

\begin{table}[t]
 \caption{Comparison on the network designs for exploring temporal dynamics from feature vectors of densely sampled frames (in accuracy \%) on the HMDB51 dataset (Split 1). We take TSN~\cite{wang2016temporal} as our backbone for extracting frame level features.}
 \fontsize{9pt}{10pt}\selectfont\centering
 \label{tab:vscnn}
 \begin{center}
   \begin{tabular}{|l|c|c|}
     \hline
     Architecture & RGB & Optical Flow \\
     \hline\hline
     TSN \cite{wang2016temporal} & 54.6 & 62.6 \\
     TSN+3-layer LSTM & 51.1 & 59.6 \\
     TSN+TSM~(AlexNet) & 51.7 & 58.0 \\
     TSN+TSM~(3-layerConvNet) & \textbf{55.0} & \textbf{63.1} \\
     \hline
   \end{tabular}
 \end{center}
\vspace{-2mm}
\end{table}

\noindent\textbf{Effectiveness of Temporal Attention.} Different frames generally have different importance levels for recognizing the action. The less relevant frames or irrelevant frames could be noise which hurts the final performance. We have designed a temporal attention module with hierarchical attentions as shown in Figure \ref{fig:shallow-cnn} (which is Figure 3 in the paper) and found that the hierarchical structure provides superior performance. Table~\ref{tab:attn} shows the comparisons with the designs having fewer attention branches. ``w Attn. ($A_0$)" denotes that only the attention $A_0$ is applied in the input (see Figure \ref{fig:shallow-cnn}).  ``w Attn. ($A_1$ \& $A_2$)" denotes the attentions $A_1$ and $A_2$ are applied after the first layer and the second layer of the shallow CNN. ``w Attn. ($A_0$ \& $A_1$ \& $A_2$)" denotes our final scheme with all the three branches of attentions included. We can that the hierarchical attention structure provides superior performance, and when our attention model is enabled (\emph{w/ Attn.}) with experiments conducted on the HMDB51 dataset (Split 1), the performance can be improved by 0.8\%, demonstrating the effectiveness of the attention mechanism. 

\begin{table*}[t]
	\caption{Accuracy (\%) of two-stream based network TSN+TSM without (w/o Attn.) and with (w/o Attn.) temporal attention on the HMDB51 dataset (Split 1).}
	\label{tab:attn}
	\fontsize{9pt}{10pt}\selectfont\centering
	\tabcolsep=14pt
	\begin{center}
		\begin{tabular}{|l|c|c|c|}
			\hline
			Model & RGB & Optical Flow & Fusion \\
			\hline\hline
			w/o Attn. & 55.0 & 63.1 & 72.4 \\
			w Attn. ($A_0$) & 54.6 & 63.3 & 72.4\\
			w Attn. ($A_1$ \& $A_2$) & 55.0 & 63.3 & 72.7 \\
			w Attn. ($A_0$ \& $A_1$ \& $A_2$)& \textbf{55.2} & \textbf{63.3} & \textbf{73.2} \\
			\hline
		\end{tabular}
	\end{center}
\end{table*}




\noindent\textbf{Influence of the Density of Temporal Sampling.} Videos of different lengths will generate \emph{VideoMaps} of different height. We can aggregate the feature vectors of dense frames of a video to form a \emph{VideoMap}. In practice, for the convenience of learning, we densely sample the video frames to have a fixed number of frames to form the \emph{VideoMap} of fixed height. In the training, we set the number of frames as 256 in considering the average length of videos. In order to measure the influence of temporal frame sampling density during testing, we have compared the performance under different temporal sampling densities, \emph{i.e.}, 25, 32, 64, 128, and 256 frames per video on the HMDB51 dataset (Split 1) and show the results in Figure \ref{fig:redun}~(a). We can see that the performance increases as the sampling density increases. This is consistent with the human perception. Our method provides a way to jointly explore the inter-frame dynamics.



In contrast, for the TSN model \cite{wang2016temporal}, the increase of the number of frames in the test will not increase the performance,  and the performance saturates at 25 frames as shown in Figure \ref{fig:redun}~(b). This is because the TSN-like approaches (\emph{e.g.}, TSN \cite{wang2016temporal}, TLE \cite{diba2017deeptemporal}) explore the temporal dynamics by simply averaging/multiplying the scores/features of the frames. The statistical information rather than the per-frame detailed information is explored for recognition. Then, more frames above a certain number of frames help little when they are enough to represent the statistical information. While, the performance of our model increases with the increase of frame density (see Figure \ref{fig:redun}~(a)).

\subsection{Comparison with the State-of-the-art }
We compare our proposed scheme with the state-of-the-art approaches for video action recognition in Table~\ref{tab:statoftheart}. We evaluate the performance on the HMDB51 and the UCF101 dataset. For both datasets, we use the provided evaluation protocol and report the mean average accuracy over the three splits. We can see our scheme achieves the best performance, with 72.7\% on the HMDB51 dataset and 94.3\%, on the UCF101 dataset. In comparison with the TSN, we achieve 4.2\% improvement on the HMDB51 dataset.

\begin{table}[h]
\caption{Performance comparisons (in accuracy \%) of our method with the other state-of-the-art methods over all the three splits.}
\label{tab:statoftheart}
 \fontsize{9pt}{10pt}\selectfont\centering
   \begin{center}
     \begin{tabular}{|l|c|c|}
       \hline
       Method & HMDB51 & UCF101 \\
       \hline\hline
       Slow Fusion CNN~\cite{karpathy2014large} & --- & 65.4 \\				
       Two-Stream CNN (VGG16)~\cite{feichtenhofer2016convolutional} & 58.5 & 91.4 \\				
       Two-Stream CNN (AlexNet)~\cite{simonyan2014two} & 59.4 & 88.0 \\
       Key Volume Mining~\cite{zhu2016key} & 63.3 & 93.1 \\

       Two-Stream CNN Fusion~\cite{feichtenhofer2016convolutional} & 65.4 & 92.5 \\
               Spatiotemporal ResNets~\cite{feichtenhofer2016spatiotemporal} & 66.4 & 93.4 \\
       \textbf{TSN (BN-Inception)}~\cite{wang2016temporal} & 68.5 & 94.0 \\			
       Spatiotemporal Multiplier Nets~\cite{feichtenhofer2017spatiotemporal} & 68.9 & 94.2 \\
       Spatiotemporal Pyramid Net~\cite{wang2017spatiotemporal} & 68.9 & 94.6 \\
               Fusion+iDT~\cite{feichtenhofer2016convolutional} & 69.2 & 93.5 \\	
       ActionVLAD~(VGG16)+iDT~\cite{Girdhar2017ActionVLAD} & 69.8 & 93.6 \\               			
       TLE:Bilinear~\cite{diba2016deep} & 71.1 & 95.6 \\
       \hline
       LRCN~\cite{donahue2015long} & --- & 82.9 \\
       C3D~\cite{tran2015learning} & --- & 85.2 \\
       C3D+iDT~\cite{tran2015learning} & --- & 90.4 \\
       C3D+LSTM~\cite{Ye2016Embedding} & 55.2 & 85.4 \\

         VideoLSTM+iDT(FV)~\cite{li2016videolstm} & 63.0 & 91.5 \\
       Multi-Granular Nets~\cite{li2016action} & 63.6 & 90.8 \\
       Multi-Stream Fusion~\cite{wu2016multi} & --- & 92.6 \\			
       Hierarchical Attention Nets~\cite{wang2016hierarchical} & 64.3 & 92.7 \\

       \hline \hline
     TSN+TSM (Ours w/o Attn.) & 72.2 & 94.1 \\
     \textbf{TSN+TSM (Ours w/ Attn.)} & 72.7 & 94.3 \\
       \hline
     \end{tabular}
   \end{center}
\vspace{-4mm}
\end{table}

Compared with the HMDB51 dataset, the accuracy improvement on the UCF101 dataset is smaller. The performance of the UCF101 dataset is approaching saturation ($>$94\%) and it becomes difficult to demonstrate the effectiveness of an approach. We will perform further study on more challenging datasets in the future.

\subsection{Visualization}

We make performance comparison for all the categories on the HMDB51 dataset to get better insights. Figure \ref{fig:histgram-gainclasses} shows the top-25 classes that our approach outperforms TSN. For some action classes such as ``Draw sword" and ``Eat", our scheme outperforms TSN even by 20\%.
In TSN, ``Draw sword'' is easy to be mistaken as ``Sword" and ``Wave" since these actions usually share some common states like  waving. Figure \ref{fig:examples} shows such an example. With our scheme capable of looking at dense frames with time order embedded rather than several sparse frames, the accuracy is improved by 23\% for the class of ``Draw sword". Similarly, ``Eat'' and ``Drink'' are prone to be confused and we find the video samples of the two classes usually have frames of similar states once the cup is away from the mouth. For the classes with time order, \emph{e.g.}, ``Stand" versus ``Sit" as shown in Figure \ref{fig:examples}, our scheme can capture the time order well thanks to the \emph{VideoMap} representation and outperforms TSN.

\begin{figure}[t]
	\begin{center}
		\includegraphics[width=1.0\linewidth]{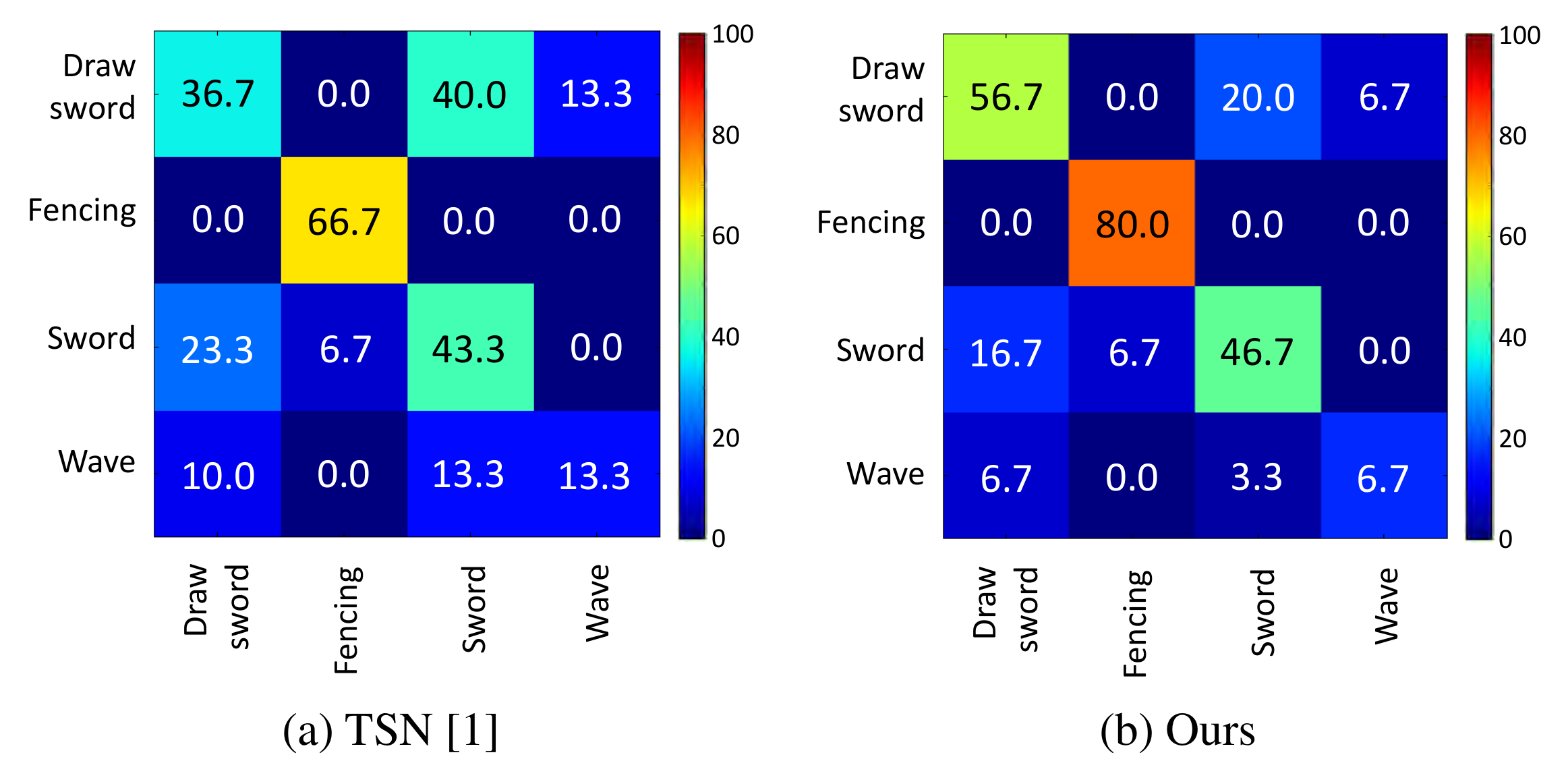}
	\end{center}
	\vspace{-0.1in}
	\caption{Comparison of partial confusion matrix with related categories of ``Draw sword", ``Fencing", ``Sword", and ``Wave". (a) TSN~[1]; (b) Ours.}
	\label{fig:cm1}
\end{figure}

\begin{figure}[t]
	\begin{center}
		\includegraphics[width=1.0\linewidth]{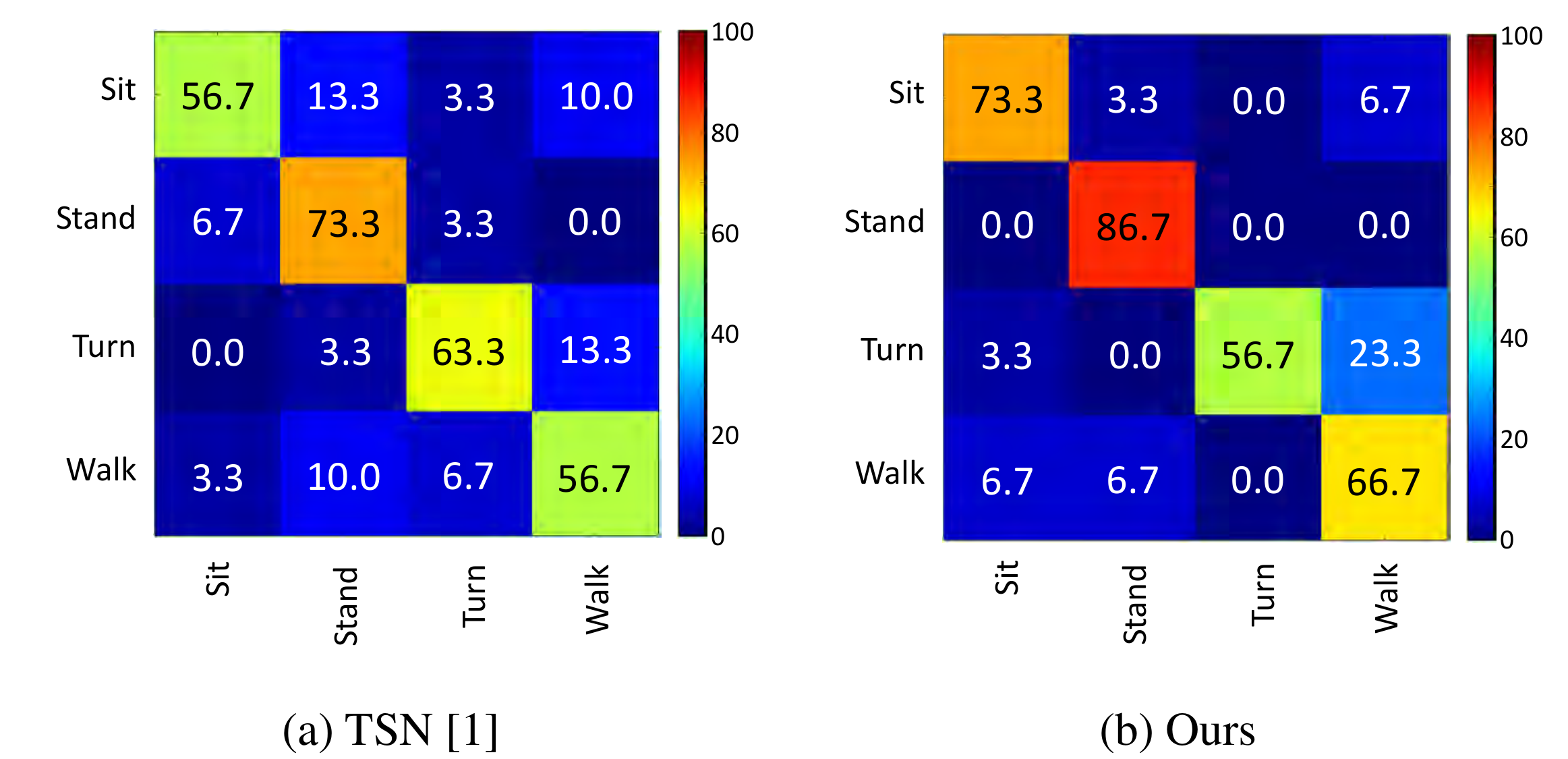}
	\end{center}
	\vspace{-0.1in}
	\caption{Comparison of partial confusion matrix with related categories of ``Sit", ``Stand", ``Turn", and ``Walk". (a) TSN~[1]; (b) Ours.}
	\label{fig:cm2}
\end{figure}

In addition, Figure \ref{fig:cm1} and Figure \ref{fig:cm2} show partial of the confusion matrix corresponding to some easily confused categories, as referred in Section 6.4. In Figure \ref{fig:cm1}, for the approach of Temporal Segment Networks (TSN)~\cite{wang2016temporal}, ``Draw sword" is easy to be mistaken as ``Sword" and ``Wave" where these actions usually share some common states like waving. Our proposed scheme performs much better than TSN since it is capable of looking at all frames and jointly making decision. In Figure \ref{fig:cm2}, the proposed TSN+TSM method can distinguish ``Sit", and ``Stand" much better since the time order information is embedded in the \emph{VideoMap} representation.

\begin{figure}[h]
 \vspace{0.1mm}
 \begin{center}
 \includegraphics[width=1.0\linewidth]{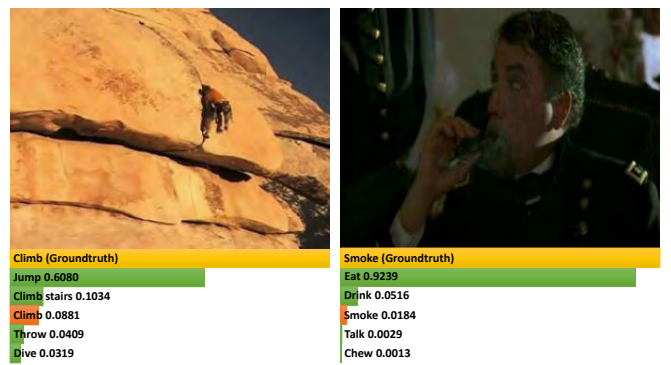}
 \caption{Examples on the failure cases of the proposed model on the HMDB51 dataset (Split 1). The main reason for these failures comes from the similar appearances in the confusing classes.}
 \label{fig:TSM-mistakes}
   \end{center}
\end{figure}

There are some failure cases of our TSM, For instance, one of action ``Climb'' is mistaken as ``Jump'', and one of ``Smoke" is mistaken as ``Eat". The video appearances in the confused classes are similar. And we show two examples in Figure~\ref{fig:TSM-mistakes}. For instance, ``Climb'' is mistaken as ``Jump", and ``Smoke" is mistaken as ``Eat". The appearances in the confused classes are similar.


Furthermore, we adopt Grad-CAM visualization technique \cite{Selvaraju2016Grad} to analyze our head ConvNet. Grad-CAM is a class-discriminative localization technique, which can provide visual explanations from the learned ConvNet model without requiring architectural changes or re-training. In Figure \ref{fig:visualization}, we can see that Grad-CAM map presents higher responses over temporal segments corresponding to the frames that persons are doing the corresponding actions. For instance, in Figure \ref{fig:visualization} (a1)--(c1), For the action of ``stand up'', there are some unrelated frames before the acting of standing up. In Figure \ref{fig:visualization} (a2)--(c2), for the action ``kick ball'', there are high responses at the first 2/3 of the temporal duration and lower responses at the remaining time duration. The temporal response characteristics are similar to that for object detection/classification, where the regions being highly correlated with the actions/objects/classes having higher responses.

\section{Conclusion}

To model the temporal-spatial evolution in video for action recognition, we propose a simple yet effective operation, Temporal-Spatial Mapping (TSM), to enable the joint analysis of the dense frames of a video. We propose a video level 2D feature representation by transforming the convolutional features of a sequence to a \emph{VideoMap}, where the temporal dynamic evolution is well embedded. We leverage a head ConvNet with temporal attention model to further explore the temporal-spatial dynamics in the VideoMap and learn effective video-level feature representation for classification. The experiment results show that the proposed scheme achieves the state-of-the-art performance, 72.7\% and 94.3\% on the HMDB51 and UCF101 dataset, respectively.


%

%

%

\ifCLASSOPTIONcaptionsoff
  \newpage
\fi



\bibliographystyle{IEEEtran}
\bibliography{reference}
\end{document}